\documentclass[10pt,twocolumn,letterpaper, cta-author]{article}

\usepackage[pagenumbers]{cvpr} % To force page numbers, e.g. for an arXiv version pagenumbers

\usepackage{graphicx}
\usepackage{caption}
\usepackage{amsmath}
\usepackage{amssymb}
\usepackage{booktabs}
\usepackage{amsmath,array,graphicx}
\usepackage{kantlipsum}
\usepackage{array}
\usepackage[dvipsnames]{xcolor}
\usepackage{colortbl}
\usepackage[accsupp]{axessibility}  % Improves PDF readability for those with disabilities.
\definecolor{myred}{cmyk}{0, 54, 62, 0}
\usepackage[pagebackref,breaklinks,colorlinks]{hyperref}

\usepackage[capitalize]{cleveref}
\crefname{section}{Sec.}{Secs.}
\Crefname{section}{Section}{Sections}
\Crefname{table}{Table}{Tables}
\crefname{table}{Tab.}{Tabs.}

\def\cvprPaperID{8905} % *** Enter the CVPR Paper ID here
\def\confName{CVPR}
\def\confYear{2023}

\begin{document}

\title{Improving Visual Representation Learning through Perceptual Understanding}

% \author{Samyakh Tukra}
% \author{Frederick Hoffman}
% \author{Ken Chatfield}
% \affil{Tractable AI}

% \author{
% Samyakh Tukra
% \and
% Frederick Hoffman
% \\[0.5cm]
% Tractable AI
% \and
% Ken Chatfield
% }

\author{
Samyakh Tukra \hspace{0.4cm} Frederick Hoffman \hspace{0.4cm} Ken Chatfield\\
Tractable AI\\
{\tt\small \{samyakh.tukra,frederick.hoffman,ken\}@tractable.ai}
}

% \author{Samyakh Tukra\\
% Tractable\\
% %Institution1 address\\
% {\tt\small samyakh.tukra@tractable.ai}
% \and
% Frederick Hoffman\\
% %Institution2\\
% %First line of institution2 address\\
% {\tt\small frederick.hoffman@tractable.ai}
% \and
% Ken Chatfield\\
% %Institution2\\
% %First line of institution2 address\\
% {\tt\small ken@tractable.ai}
% }
\maketitle

%%%%%%%%% ABSTRACT
\begin{abstract}
   We present an extension to masked autoencoders (MAE) which improves on the representations learnt by the model by explicitly encouraging the learning of higher scene-level features. We do this by: (i) the introduction of a perceptual similarity term between generated and real images (ii) incorporating several techniques from the adversarial training literature including multi-scale training and adaptive discriminator augmentation. The combination of these results in not only better pixel reconstruction but also representations which appear to capture better higher-level details within images. More consequentially, we show how our method, Perceptual MAE, leads to better performance when used for downstream tasks outperforming previous methods. We achieve $78.1\%$ top-1 accuracy linear probing on ImageNet-1K and up to $88.1\%$ when fine-tuning, with similar results for other downstream tasks, all without use of additional pre-trained models or data.
\end{abstract}

%%%%%%%%% BODY TEXT
\section{Introduction}
\label{sec:intro}
Self-supervision provides a powerful framework for training deep neural networks without relying on explicit supervision or labels where learning proceeds by predicting one part of the input data from another. Approaches based on denoising autoencoders~\cite{Vincent2008}, where the input is masked and the missing parts reconstructed, have shown to be effective for pre-training in NLP with BERT~\cite{47751}, and more recently similar techniques have been applied for learning visual representations from images~\cite{pmlr-v119-chen20s, 50650, MaskedAutoencoders2021, bao2022beit}. Such methods effectively use image reconstruction as a \textit{pretext task} on the basis that by learning to predict missing patches useful representations can be learnt for downstream tasks.

One challenge when applying such techniques to images is that, unlike language where words contain some level of semantic meaning by design, the pixels in images are natural signals containing high-frequency variations. Therefore, image-based denoising autoencoders have been adapted to avoid learning trivial solutions to reconstruction based on local textures or patterns. BEiT~\cite{bao2022beit} uses an intermediary codebook of patches such that pixels are not reconstructed directly, whilst MAE~\cite{MaskedAutoencoders2021} masks a high proportion of the image to force the model to learn how to reconstruct whole scenes with limited context.

In this paper, we build upon MAE and ask how we can move beyond the implicit conditioning of high masking ratios to explicitly incorporate the learning of higher-order `semantic' features into the learning objective. To do this, we focus on introducing scene-level information by adding a perceptual loss term~\cite{Johnson2016Perceptual}. This works by constraining feature map similarity with a second pre-trained network, a technique which has been shown empirically in the generative modelling literature to improve perceptual reconstruction quality~\cite{zhang2018perceptual}. In addition, this also provides a mechanism to incorporate relevant scene-level cues contained in the second network (which could be \emph{e.g.}\ a strong ImageNet classifier or a pre-trained language-vision embedding).

One of the benefits of MAE is that it can rapidly learn strong representations using only self-supervision from the images in the target pre-training set. To maintain this property, we introduce a second idea: tying the features not with a separate network, but with an adversarial discriminator trained in parallel to distinguish between real and generated images. Both ideas combined result in not only a lower reconstruction error, but also learnt representations which better capture details of the scene layout and object boundaries (see Figure~\ref{fig:attn_maps}) without either explicit supervision or the use of hand-engineered inductive biases.

Finally, we build on these results and show that techniques from the generative modelling literature such as multi-scale gradients~\cite{progressive_gans} and adaptive discriminator augmentation~\cite{Karras2020ada} can lead to further improvements in the learnt representation, and this also translates into a further boost in performance across downstream tasks. We hypothesise that the issues that these methods were designed to overcome, such as mode collapse during adversarial training and incomplete learning of the underlying data distribution, are related to overfitting on low-level image features.

Our contributions can be summarized as follows: (i)~we introduce a simple and self-contained technique to improve the representations learnt by masked image modelling based on perceptual loss and adversarial learning, (ii)~we perform a rigorous evaluation of this method and variants, and set new state-of-the-art for masked image modelling without additional data for classification on ImageNet-1K, object detection on MS COCO and semantic segmentation on ADE20K, (iii)~we demonstrate this approach can further draw on powerful pre-trained models when available, resulting in a further boost in performance and (iv)~we show our approach leads qualitatively to more `object-centric' representations and stronger performance with frozen features (in the linear probe setting) compared to MAE.
%------------------------------------------------------------------------
\section{Related Work}
\label{sec:related_work}

\subsection{Masked Image Modelling}

Self-supervised learning has led to learning systems that do not depend on data labelling, where the raw data itself provides the supervisory signal for training. This results in models with feature representations that are generalisable to many tasks. Self-supervised learning has shown considerable success in Natural Language Processing (NLP)~\cite{NIPS2017_3f5ee243, NEURIPS2020_1457c0d6, liu2019roberta}%, devlin2018pretraining, conneau-etal-2020-unsupervised}
, where random parts of the input text are masked and the model is tasked with predicting the invisible content. This has become the de facto method of pre-training NLP models. Compared to this direct-prediction approach, the first performant approaches to self-supervised visual representation learning instead used predefined discriminative tasks such as estimating distortions of the input image~\cite{NIPS2014_07563a3f, Gidaris2018UnsupervisedRL}, patch re-ordering~\cite{7410524, 10.1007/978-3-319-46466-4_5}, re-coloring a grayscale image input~\cite{zhang2016colorful}, and contrastive learning~\cite{Oord2018RepresentationLW, pmlr-v119-chen20j}.

Recently, inspired by NLP and facilitated by the advent of Transformer models (Vision Transformers~\cite{50650} in particular) masked image modelling (MIM) returns to the idea of direct-prediction, randomly masking pixels of an input image before predicting the invisible content. Early work included iGPT~\cite{pmlr-v119-chen20s} which downsized images and then directly predicted unknown pixel values in an autoregressive manner. Recent methods have moved towards predicting full resolution patches in an autoencoder configuration~\cite{bao2022beit, MaskedAutoencoders2021, Xie_2022_CVPR}. BEiT~\cite{bao2022beit} relies on an additional generative model (\textit{dVAE}~\cite{pmlr-v139-ramesh21a}) pre-trained on a large corpus of images (250M) with the pretext task to predict for masked matches the closest visual token from a pre-trained codebook. MAE~\cite{MaskedAutoencoders2021} takes a simpler approach, demonstrating that direct pixel prediction of masked regions using Mean Squared Error is also effective when a very large proportion of the image is masked out.

These recent methods produce strong performance when fine-tuning over downstream tasks, but generally discriminative self-supervision has continued to be more performant in a linear probe setting~\cite{MaskedAutoencoders2021} suggesting focusing on the learning of features of the right level of abstraction remains a challenge. We seek to address this in our work.

\subsection{Perceptual Similarity} \label{sec:perceptual}

The aim of perceptual similarity~\cite{zhang2018perceptual} is to mimic human visual perception. Humans are capable of understanding images on an abstract level, relying on high level concepts and semantic cues that define the underlying relationships between different entities in the frame. Perceptual similarity aims to mimic this human-like judgement by defining metrics which encode \textit{perceptual distance}, with this being higher for image representations that similarly better capture visual semantic concepts.

Structural Similarity Index (SSIM), an early form of perceptual loss, attempts to capture properties of an image that when varied are perceived by humans as substantially different. Images are compared on three key features: (i) luminance (\textit{i.e.}\ pixel intensity), (ii) contrast and (iii) structure~\cite{journals/tip/WangBSS04}. Further work extended this to compute similarity at multiple scales~\cite{1292216}. In parallel other similar metrics have been proposed such as: Peak Signal to Noise ratio (PSNR)~\cite{5596999}, Feature Similarity Index (FSIM)~\cite{5705575} and HDR-VDP-2~\cite{10.1145/2010324.1964935}.

An alternative approach to capturing perceptual similarity is to not compare differences between pixels but instead compute the differences between the intermediary features learnt by a neural network and those extracted from a second fixed network pre-trained in a supervised manner on a large dataset, on the basis that these capture the higher-level semantically meaningful features required for accurate classification. This is the approach taken by~\cite{Johnson2016Perceptual} who pre-train a VGG network on ImageNet and then use this for learning. Such a \textit{feature matching} based approach has been successfully applied to many tasks since in computer vision~\cite{9352486, wang2018pix2pixHD, wang2018esrgan}.% In our work, we draw on the feature matching based approach to perceptual similarity but remove the requirement for a pre-trained network, instead learning perceptual similarity dynamically.

In our work, we draw on the feature matching based approach to perceptual similarity but remove the requirement for a pre-trained network, instead learning perceptual similarity dynamically. In parallel to this work, PeCo~\cite{dong2021peco} also experiments with perceptual loss to prepare a perceptually aware codebook for masked image modelling. However, in our case we apply this directly during pre-training which is much more effective, as it enables the encoder to learn directly higher-level cues from the second network.

\subsection{Generative modelling}

If perceptual similarity tries to capture the semantic structure of images, generative models aim to capture the underlying distribution of the image data. An example is Generative Adversarial Networks (GANs)~\cite{karnewar2019msg, Karras2020ada, progressive_gans, 9156570}. The samples created by a generator model are evaluated by a separately trained discriminator model which is tasked with determining real images from generated images. This is trained in parallel with the generator using an adversarial loss function. Since GANs learn the underlying data distribution implicitly via a discriminator the original formulation~\cite{NIPS2014_5ca3e9b1} could produce high-fidelity images but suffered from training instability and mode collapse, where the network was only able to capture a subset of the variance present in the data distribution. Subsequent work including Pro-GAN~\cite{progressive_gans} and MSG-GAN~\cite{karnewar2019msg} introduced the idea of generation at multiple scales to stabilise the generator, enabling a more complete capture of the underlying data distribution.

The StyleGAN family of papers~\cite{8953766, 9156570, Karras2020ada} introduced several improvements to the learning of the discriminator further designed to improve the stability of generated images. Perceptual path length regularisation~\cite{zhang2018perceptual} enforces that small changes in the input latent code (in our work: the input to the decoder) lead to changes of a similar magnitude in the feature maps of the discriminator, thus ensuring good normalisation of the input codes. Adaptive Discriminator Augmentation (ADA)~\cite{Karras2020ada} enables the use of heavy augmentation when training the discriminator, avoiding overfitting even with smaller volumes of training data, whilst ensuring these augmentations do not affect the output of the generator. Both encourage the underlying feature space to be more stable to small changes in low-level image statistics. In this work, we explore if these additions, along with multi-scale learning described above additionally incorporated into the masked autoencoder architecture, can therefore help to learn richer, high-level representations when using adversarial training for masked image modelling.

An alternative to GANs for generative modelling is provided by explicit generative models which aim to capture the underlying data distribution directly. These methods avoid some of the issues such as mode collapse suffered by implicit modelling but, given the need to model the full distribution, generally are more sensitive to the volume of data used for training. Examples include Variational Auto-Encoders (VAE)~\cite{Kingma2014, higgins2017betavae}, Flow-based models~\cite{DBLP:conf/iclr/DinhSB17, NEURIPS2018_d139db6a} and Diffusion models~\cite{NEURIPS2021_49ad23d1, pmlr-v162-nichol22a}. %VAEs are a variant of standard autoencoders that impose a probabilistic prior on the learnt latent space. A separate surrogate loss (Kullback-Leibler divergence) is then used to align the learnt latent distribution with the prior, capturing an approximation of the underlying data distribution. 
VQ-VAE~\cite{NIPS2017_7a98af17} builds on VAE by learning a discrete latent space rather than a continuous one by the creation of a codebook. Recently this was applied for masked image modelling in BEiT~\cite{bao2022beit}, where the rich latent representations learnt by a VQ-VAE model pre-trained on large data~\cite{pmlr-v139-ramesh21a} are used as a prediction target. In this work, we also explore using such a large pre-trained VQ-VAE model when combined with perceptual similarity loss.%Flow-based models are closely related to VAEs and learn the distribution explicitly via normalising flows \textit{i.e.}~a sequence of invertible transformations. Diffusion models learn to destroy inputs with successive noise addition whilst simultaneously learning to reconstruct the original image during the backward pass.

%Explicit generative models are applied for masked image modelling in the BEiT paper~\cite{bao2022beit}, where the rich latent representations learnt by a VQ-VAE model pre-trained on large data are used as a prediction target. In this work, we also explore using such a large pre-trained VQ-VAE model combined with perceptual similarity loss.

% ------------------------------------------------------------------------
\section{Methodology}

The learning framework used for this work is based on MAE~\cite{MaskedAutoencoders2021}. In Section~\ref{perceptual_loss} we describe how the MAE loss is extended with a perceptual loss term. In Section~\ref{adversarial_losses} we then describe variants of adversarial loss which is also added to the objective. In Section~\ref{msg_mae_model} we describe modifications to the MAE architecture to maximize learning in the encoder stage when using multi-scale gradients.

\subsection{MAE with Perceptual Loss}
\label{perceptual_loss}
The pixel reconstruction loss from the original MAE formulation is extended to include a perceptual loss term:

\begin{equation}
  L^G = ||G(I_{m}) - I||_{1} + L_{perceptual}^{G}
\end{equation}

Where $G$ is the MAE model, $I$ is the original image and $I_{m}$ is the original image randomly masked. We follow the convention in the generative modelling literature, and use L1 loss rather than L2 loss for the reconstruction term.

\textbf{MS-SSIM:} Our baseline perceptual loss is based on structural similarity index, specifically the multi-scale variant (MS-SSIM)~\cite{1292216}. The multi-scale component aids in reducing artefacts formed around the edges of the output reconstructed image $I'$. The perceptual loss term is thus:

\begin{equation}
  L_{ssim}^{G}= \frac{1}{N}\sum_{i,j}\alpha\frac{1-SSIM(G(I_{m})_{ij},I_{ij})}{2}
\end{equation}

Where $i, j$ are the pixel indexes and $N$ is the total number of scales, set to 4. We use a $3 \times 3$ block filter for each scale. $\alpha$ is a weighting constant, with the L1 error weighted by the inverse $(1 - \alpha)$.

\textbf{Feature matching:} Based on the feature and style reconstruction losses of~\cite{Johnson2016Perceptual} our second perceptual loss relies on a separate \textit{loss network} with the decoder network encouraged to have similar feature representations as each corresponding layer of the loss network $\phi$.

In the original formulation $\phi$ is a fixed VGG network pre-trained on ImageNet as described in Section~\ref{sec:perceptual}. To avoid this dependency on an external pre-trained network, we instead introduce an additional discriminator network $D$ which will act as our loss network $\phi$. This is trained in an adversarial setup to distinguish between the reconstructed image from the decoder $G$ and the original image prior to masking. The intuition is that the features learnt through this task also contain higher-order perceptual cues which can be used to guide training of the decoder.
% This is trained in a GAN-style training setup with the decoder acting as the generator $G$.

The perceptual loss comprises two parts. The first transfers high-level semantics (individual features are similar), with a second style term added which learns overall image statistics (correlations between features across the image are similar) giving:

% In addition to penalising differences in feature activations via the feature reconstruction loss, we find that additionally penalising differences in the correlation of feature activations within each image via the style reconstruction loss (which ties together overall color and texture) further stabilises the training. The perceptual loss then becomes:

\begin{equation}
\label{feat_matching}
\begin{split}
    L_{feat}^{G} = & \delta_{f} \sum_{j=1}^{J}\frac{1}{N_{j}}[||\phi^{j}(G(I_{m}))-\phi^{j}(I)||_1] + \\
    & \delta_{s} \sum_{j=1}^{J}\frac{1}{N_{j}}[||\Psi(\phi^{j}(G(I_{m})))-\Psi(\phi^{j}(I))||_1]
\end{split}
\end{equation}

Where $j$ is the index of the layer, $N_j$ denotes the number of elements in each layer, $\Psi$ is the Gram matrix function~\cite{7780634} and $\delta_{f}$ and $\delta_{s}$ are constant weighting factors.

Additionally, the adversarial loss is added to $L^{G}$. Any adversarial loss function can be used, but for our baseline experiments we use \textit{LS-GAN}~\cite{Mao2017LeastSG} which has shown to achieve more stable optimisation over the original min-max classification loss. This gives us the generator-discriminator loss pair:

\begin{equation}
    L_{adv}^{D}= \frac{1}{2}[(D(I)-1)^{2}] + [(D(G(I_{m}))^{2}]
\end{equation}
\begin{equation}
    L_{adv}^{G}= \frac{1}{2}[(D(G(I_{m}))-1)^{2}]
\end{equation}

The full loss function for the decoder then becomes:

\begin{equation}
  L^G = ||G(I_{m}) - I||_{1} + L_{feat}^{G} + L_{adv}^{G}
\end{equation}

Beyond the feature matching acting as a learnt perceptual loss, adding this term also has the further advantage of stabilising adversarial training.

\textbf{dVAE perceptual:} To provide a perceptual learning baseline also using the stronger supervision from a pre-trained network for comparison, we experiment with feature matching loss with the discrete variational autoencoder (dVAE) from~\cite{pmlr-v139-ramesh21a}. For this, the same feature matching loss in Equation~\ref{feat_matching} above is used with the pre-trained dVAE encoder model component acting as the loss network $\phi$. This then is the perceptual loss term, giving the full loss function for the decoder:

\begin{equation}
  L^G = ||G(I_{m}) - I||_{1} + L_{feat}^{G}
\end{equation}

The dVAE was trained for image tokenization on the DALL-E dataset, comprising 250 million images. Therefore, the rich features encoded in its weights provide strong higher-order perceptual cues for decoder training.

\subsection{Adversarial Training Variants}
\label{adversarial_losses}
For feature matching based perceptual learning, any adversarial loss function can be used. In addition to the LS-GAN loss used in our baseline model, we experimented with two further variants, introduced below. Both were formulated to address issues with the original GAN formulation such as training instability and mode collapse~\cite{karnewar2019msg}. We hypothesize that the richer distributions learnt by these methods will provide stronger cues for perceptual learning.

\textbf{MSG-GAN: } To stabilise the training of the generator, MSG-GAN~\cite{karnewar2019msg} allows for the flow of gradients from the discriminator to the generator at multiple scales. This is done by adding skip connections from intermediate layers of the generator to intermediate layers of the discriminator. The loss function for training $D$ and $G$ remains unchanged.

\textbf{StyleGANv2-ADA: } We take all modifications made to the discriminator in the StyleGANv2-ADA~\cite{Karras2020ada} paper. Building on MSG-GAN, perceptual path regularisation between the decoder input and discriminator feature maps is added. Adaptive discriminator augmentation is also applied to all samples during training. The loss function for training $D$ and $G$ remains unchanged.

\subsection{Model Architecture}
\label{msg_mae_model}

\begin{figure*}[t]
    \centering
    \includegraphics[width=1.0\textwidth]{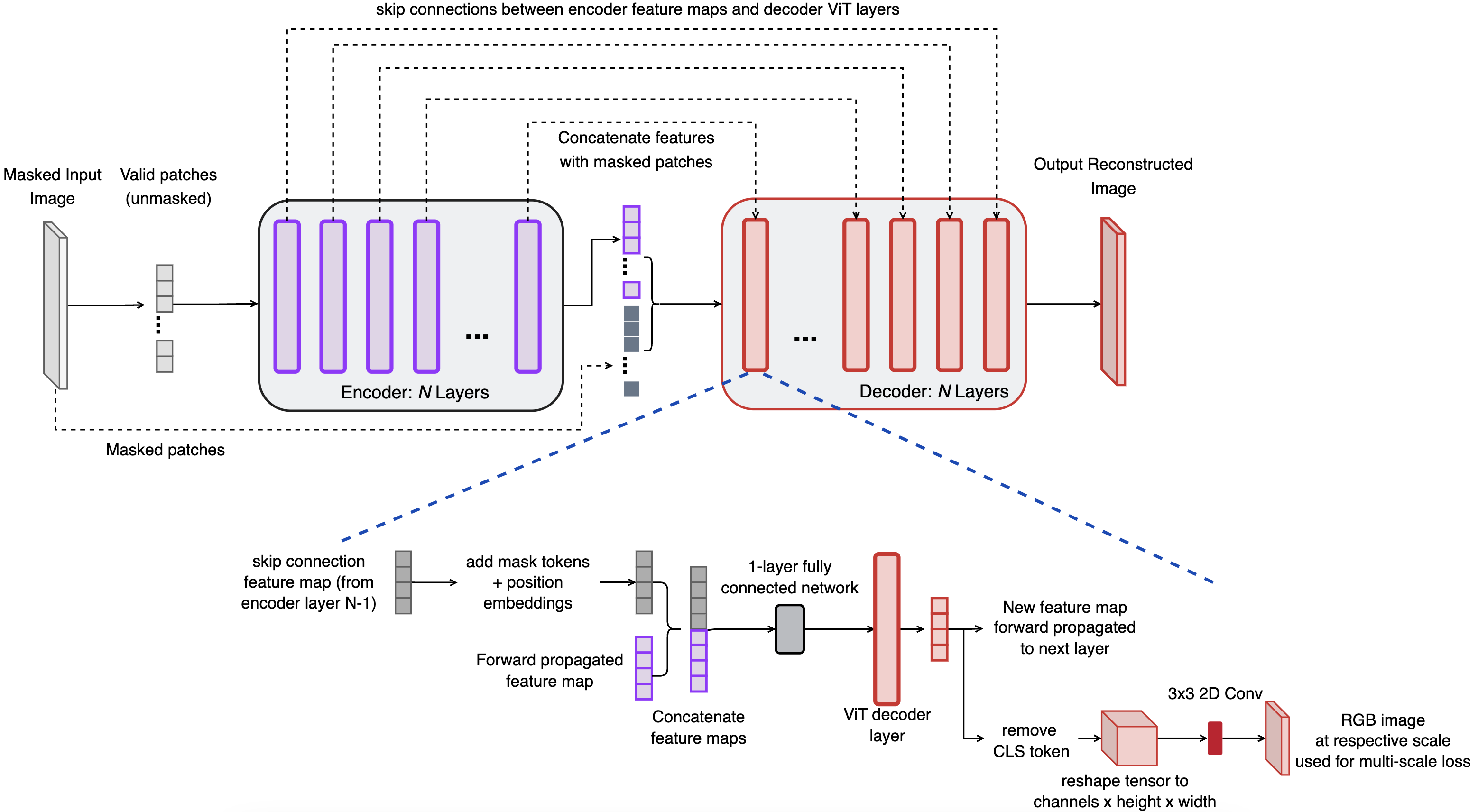}
    \caption{The Multi-Scale Gradient MAE (\textit{MSG-MAE}) architecture. Dotted lines denote skip connections and solid forward propagation.}
    \label{fig:msg_mae}
\end{figure*}

One issue with the multi-scale GAN formulation used for both \textit{MSG-GAN} and \textit{StyleGANv2-ADA} methods is that the multi-scale learning occurs via skip connections between the discriminator $D$ and decoder $G$. This means that only the decoder benefits from the multi-scale gradient penalty during training, and this is removed post pre-training, leaving only the encoder when adapting to downstream tasks.

To distribute the learning more evenly between encoder and decoder, similar to U-Net~\cite{RFB15a} we additionally introduce skip connections between intermediate encoder and decoder layers, as shown in Figure~\ref{fig:msg_mae}. In this modified architecture, which we term \textbf{MSG-MAE}, multi-scale signal is shared also with the encoder.
In further detail: mask tokens are added to the encoder feature maps of dimension $d$ along with positional embeddings. Via the skip connections, this is concatenated with the decoder feature map of the matching scale, creating a sequence of length $2d$. This combined feature is passed through a single learnt linear layer to return the dimension to $d$. The output is then processed by the decoder transformer layer, with the result forward propagated to the next layer and simultaneously converted to an RGB image at the required scale for the multi-scale loss.

\section{Implementation Details}

\textbf{Pre-training:} We use the ViT-B and ViT-L architectures from the MAE paper~\cite{MaskedAutoencoders2021}, with ViT-B and ViT-L trained for 300 and 1600 epochs respectively over the ImageNet-1K (IN1K)~\cite{5206848} training set. In each case, the input patch size is fixed to 16x16 and we mask 75\% of input patches during training. %For the ViT-B encoder, the width is 768 dimensions and comprises 12 layers each with 12 self-attention heads. For the ViT-L encoder, the width is 1024 dimensions and comprises 24 layers each with 16 self-attention heads. 
For both ViT-B and ViT-L, the decoder architecture remains consistent. %The decoder comprises 8 layers each with 16 self-attention heads and a width of 512 dimensions. 
The model dimensions, hyperparameters and data augmentation strategies follow those of the original MAE paper~\cite{MaskedAutoencoders2021} and we train with a batch size of 16. The Adam optimizer is used with a weighted decay, where the learning rate is 0.00015, weight decay is 0.05 (cosine strategy), 40 warm-up epochs are used and the momentum parameters $\beta_{1}$ and $\beta_{2}$ are 0.9 and 0.95.

In our experiments, the weighting factors in $L_{feat}^{G}$ (where applied) is given a weighting factor $\delta_{f}$ of 0.05. The $L_{ssim}^{G}$ weighting factor $\alpha$ is set to 0.85. In both cases, the result is to focus learning on the perceptual term, with the smaller $\delta_{s}$ value still resulting in a large weighting once accounting for the larger relative magnitude of the $L_{feat}^{G}$ term. The parameter choice is based on other works in the literature~\cite{9352486, Liu2018ImageIF, wang2018pix2pixHD}. $\delta_{s}$ is given a fixed value of 40. In order to give time for the discriminator to learn new features with which to compute perceptual similarity, a training schedule whereby the perceptual loss term $L_{G}$ is applied only on even numbered epochs is used. This avoids the generator distribution collapsing to that of the discriminator and ensures well-balanced learning. All experiments were conducted on a GPU cluster consisting of 8xV100 Nvidia GPUs.

\textbf{Fine-tuning:} For fine-tuning, we take the pre-trained MAE encoder model and replace the decoder architecture with a task-specific head (initialised with random weights), similar to~\cite{MaskedAutoencoders2021}. For image classification we replace the decoder with the original ViT model classification head~\cite{50650}. For object detection and segmentation on MS-COCO~\cite{502} we use a Mask-RCNN~\cite{8237584} decoder model and for ADE-20K semantic segmentation~\cite{zhou2019semantic} we adopt the UperNet model~\cite{xiao2018unified} as our decoder. When transferring to downstream tasks the intermediate feature maps of MSG-MAE are not used for the classification tasks. However, they are used for detection and segmentation, given the Mask-RCNN and UperNet architectures operate at multiple scales. %(shown in Figure~\ref{fig:downstream} A). 
%The Mask-RCNN backbone is based on Feature Pyramid Networks (FPNs)~\cite{8099589} and we adapt the ViT encoder model accordingly. Specifically, as the encoder is composed of multiple ViT transformer layers outputting feature maps at single scale (unlike convolutional layers), we extract feature maps at 4 layer intervals i.e. $[0,4,8,12]$. Feature maps are resampled to the respective size required by original Mask-RCNN head. For Upsampling bi-linear interpolation is used with a scale factor of 2 (if required) followed by a 3x3 convolution. For downsampling the features are reshaped to a square matrix followed by a 3x3 convolution. For segmentation on ADE20K~\cite{zhou2019semantic} we adopt the UperNet model~\cite{xiao2018unified} as our decoder. This follows similar strategy as Mask-RCNN, necessitating a FPN backbone and the same processing steps described above are applied.% as illustrated in Figure~\ref{fig:downstream} B. 
All fine-tuned models are trained with an Adam optimizer with weighted decay. The learning rate is 0.001, weight decay is 0.05 (cosine strategy), warm-up epochs 5, and momentum parameters $\beta_{1}$ and $\beta_{2}$ are 0.9 and 0.95.

%\begin{figure*}[t]
%    \centering
%    \includegraphics[width=0.85\textwidth]{downstream.png}
%    \caption{Showing the augmentation performed on the MAE pre-trained encoder architecture for downstream fine-tuning. A: the architecture for object detection / segmentation on MS-COCO; B: the architecture for segmentation on ADE20K.}
%    \label{fig:downstream}
%\end{figure*}

%------------------------------------------------------------------------
\section{Experiments}
\label{results}

In this section, we evaluate our models following self-supervised pre-training on the ImageNet-1K (IN1K)~\cite{5206848} training set. We first explore the main properties of the learnt representations in Section~\ref{main_properties} in terms of (i) the fidelity of reconstructed output, (ii) the qualitative attention maps from the pre-trained model and (iii) linear probe results for downstream classification. Following this, in Section~\ref{transfer_results} we show the downstream performance of our models for transfer learning comparing this to previous work: fine-tuning on ImageNet-1K for classification, COCO for object detection and ADE20K for segmentation.

\begin{table}
\caption{Image reconstruction quality evaluation on ImageNet-1K. The ViT-B architecture is used. For columns with red headers, lower value is better and for the columns with green headers, higher value is better. The best result is highlighted in bold. Methods with $\dagger$ use the original MAE architecture, otherwise MSG-MAE is used.}
  \centering
  \small
  \begin{tabular}{>{\kern-\tabcolsep}lccccc<{\kern-\tabcolsep}}
    \toprule
    Loss Function & \cellcolor{pink} L1 & \cellcolor{lime} PSNR & \cellcolor{lime} SSIM & \cellcolor{lime} IS & \cellcolor{pink} FID \\ [0.5ex] 
     \midrule
     MSE $\dagger$ & 0.25 & 0.38 & 0.76 & 6.33 & 42.7 \\
     MS-SSIM + L1 $\dagger$ & 0.21 & 0.41 & 0.82 & 8.01 & 31.6 \\
     LS-GAN-P $\dagger$ & 0.16 & 0.53 & 0.92 & 16.2 & 28.2 \\
     MSG-GAN-P & 0.11 & 0.55 & \textbf{0.94} & 32.1 & 19.0 \\
     \textbf{StyleGANv2-ADA-P} & \textbf{0.06} & \textbf{0.58} & 0.91 & \textbf{36.8} & \textbf{10.3}\\
    \bottomrule
  \end{tabular}
  \label{tab:table1}
\end{table}

\begin{figure*}[t]
    \centering
    \includegraphics[width=1.0\textwidth]{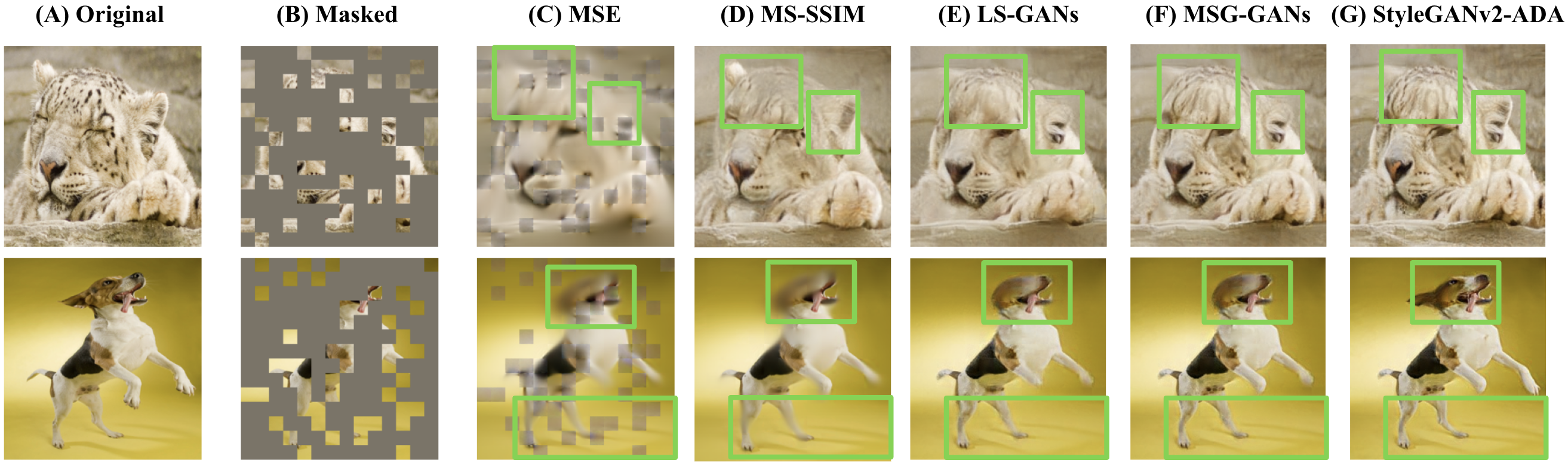}
    \caption{Differences in reconstruction quality of the model variants on samples from the ImageNet-1K validation set. The key areas of focus are highlighted in green. Columns (A-B) show the ground truth image and masking, (C-G) the reconstructed image for each method.}
    \label{fig:recon_focus}
\end{figure*}

% \begin{table*}[p]
% \caption{Image reconstruction quality evaluation on ImageNet-1K. The ViT-B architecture is used. For columns with red headers, lower value is better and for the columns with green headers, higher value is better.  The best result is highlighted in bold.}
%   \centering
%   \small
%   \begin{tabular}{lccccc}
%     \toprule
%     Loss Function & \cellcolor{pink} L1 & \cellcolor{lime} PSNR & \cellcolor{lime} SSIM & \cellcolor{lime} IS & \cellcolor{pink} FID \\ [0.5ex] 
%      \midrule
%      MSE & 0.25 & 0.38 & 0.76 & 6.33 & 42.7 \\
%      MS-SSIM + L1 & 0.21 & 0.41 & 0.82 & 8.01 & 31.6 \\
%      LS-GAN-P & 0.16 & 0.53 & 0.92 & 16.2 & 28.2 \\
%      MSG-GAN-P & 0.11 & 0.55 & \textbf{0.94} & 32.1 & 19.0 \\
%      \textbf{StyleGANv2-ADA-P} & \textbf{0.06} & \textbf{0.58} & 0.91 & \textbf{36.8} & \textbf{10.3}\\
%     \bottomrule
%   \end{tabular}
%   \label{tab:table1}
% \end{table*}

\begin{table*}[t]
\caption{\textbf{Classification performance on ImageNet1K} (IN1K). All models are pre-trained via self-supervision followed by either training of a linear probe or fine-tuning. Loss functions with `-P' include a perceptual loss term. The best result is highlighted in bold.}
%\small
\begin{center}
 \begin{tabular}{c c c c c c} 
 \toprule
 \multicolumn{3}{c}{} & Linear probe & \multicolumn{2}{c}{Fine-tuning}\\
 Method & Loss Function & Pre-training Data & ViT-B & ViT-B & ViT-L \\ [0.5ex]
 \midrule
 iGPT \cite{pmlr-v119-chen20s} & Cross Entropy & IN1K & -- & 66.5 & -- \\
 DINO \cite{caron2021emerging} & Cross Entropy & IN1K & \textbf{78.2} & 82.8 & -- \\
 MoCo v3 \cite{chen2021mocov3} & InfoNCE \cite{Oord2018RepresentationLW} & IN1K & 76.7 & 83.2 & 84.1 \\
 \addlinespace
 BEiT \cite{bao2022beit}& Negative Log Likelihood & IN1K + DALL-E & 56.7 & 83.2 & 85.2 \\
 MAE \cite{MaskedAutoencoders2021} & MSE & IN1K & 67.8 & \hspace{0.5em}83.6\textsuperscript{ a} & 85.9 \\ 
 MAE & MS-SSIM + L1 & IN1K & 71.2 & 84.1 & 86.3 \\
 MAE & LS-GAN-P & IN1K & 72.5 & 84.5 & 86.5 \\
 \addlinespace
 MSG-MAE & MSG-GAN-P & IN1K & 75.6 & \hspace{0.5em}85.3\textsuperscript{ b} & 87.2 \\
 \textbf{MSG-MAE} & \textbf{StyleGANv2-ADA-P} & IN1K & \textbf{78.1} & \textbf{86.2} & \textbf{88.1}\\
 \midrule
 {MAE} & {dVAE-P} & {IN1K + DALL-E} & {\textbf{79.8}} & {\textbf{86.9}} & {\textbf{88.6}} \\
 \midrule
 \textit{MAE} & \textit{LS-GAN} & \textit{IN1K} & -- & \textit{83.3} & \textit{85.3} \\
 \textit{MSG-MAE} & \textit{MSG-GAN} & \textit{IN1K} & -- & \hspace{0.5em}\textit{84.7}\textsuperscript{ c} & \textit{86.5} \\
 \addlinespace
 \textit{MAE} & \textit{MSG-GAN-P} & \textit{IN1K} & -- & \hspace{0.5em}\textit{83.2}\textsuperscript{ d} & \textit{85.6} \\
 \textit{MAE} & \textit{StyleGANv2-ADA-P} & \textit{IN1K} & -- & \textit{84.5} & \textit{86.2} \\
 % \hline
 % MAE & dVAE-P & IN1K + DALL-E & -- & -- & -- \\
 % \addlinespace
 % \textit{MAE} & \textit{dVAE-P} & \textit{IN1K + DALL-E} & \textit{\textbf{79.8}} & \textit{\textbf{86.9}} & \textit{\textbf{88.6}} \\
 \bottomrule
 
\end{tabular}
\end{center}
\label{tab:imagenet1k}
\end{table*}

\begin{table*}[h]
\caption{\textbf{Object detection and semantic segmentation performance on MS COCO and ADE20K.} All models were pre-trained using the ImageNet-1K training set (without labels). Loss functions with `-P' include a perceptual loss term. Best results are highlighted in bold.}
%\small
\begin{center}
 \begin{tabular}{c c c c c c} 
 \toprule
 \multicolumn{3}{c}{} & \multicolumn{2}{c}{MS COCO} & \multicolumn{1}{c}{ADE20K}\\
 Method & Loss Function & Pre-training Data & mAP \textit{Box} & mAP \textit{Mask} & mIoU\\ [0.5ex]
 \midrule
 DINO \cite{caron2021emerging}& Cross Entropy & IN1K & -- & -- & 44.1 \\
 MoCo v3 \cite{chen2021mocov3}& InfoNCE \cite{Oord2018RepresentationLW} & IN1K & 47.9 & 42.7 & 47.3 \\
 \addlinespace
 BEiT \cite{bao2022beit}& Negative Log Likelihood & IN1K + DALL-E & 49.8 & 44.4 & 47.1 \\
 MAE \cite{MaskedAutoencoders2021} & MSE & IN1K & 50.3 & 44.9 & 48.1 \\
 MAE & MS-SSIM + L1 & IN1K & 50.8 & 45.1 & 48.8 \\
 MAE & LS-GAN-P & IN1K & 51.4 & 45.4 & 49.2 \\
 \addlinespace
 MSG-MAE & MSG-GAN-P & IN1K & 52.3 & 45.8 & \hspace{0.5em}49.7\textsuperscript{ b} \\
 \textbf{MSG-MAE} & \textbf{StyleGANv2-ADA-P} & IN1K & \textbf{53.5} & \textbf{46.1} & \textbf{50.4} \\
 \midrule
 {MAE} & {dVAE-P} & {IN1K + DALL-E} &  {\textbf{53.9}} & {\textbf{46.4}} & {\textbf{50.9}} \\
 \midrule
 \textit{MAE} & \textit{MSG-GAN-P} & \textit{IN1K} & \textit{50.9} & \textit{45.9} & \hspace{0.5em}\textit{49.3}\textsuperscript{ d} \\
 \textit{MAE} & \textit{StyleGANv2-ADA-P} & \textit{IN1K} & \textit{51.8} & \textit{45.5} & \textit{49.1} \\
 % \hline
 % MAE & dVAE-P & IN1K & -- \\
 \bottomrule
\end{tabular}
\end{center}
\label{tab:coco_ade20k}
\end{table*}

\subsection{Main properties}
\label{main_properties}

\textbf{Image reconstruction.} In Table~\ref{tab:table1} we evaluate the reconstruction quality of the decoder stage of our models over the IN1K validation set using the following quantitative measures: L1 error, Peak Signal to Noise Ration (PSNR), Structural Similarity Index (SSIM)~\cite{1292216}, Inception Score (IS)~\cite{NIPS2016_8a3363ab} and Fréchet inception distance (FID)~\cite{10.5555/3295222.3295408}. These experiments use the ViT-B variant of the encoder, pre-trained using each of our perceptual losses.

For each of our methods, we observe a gradual increase in the fidelity of the reconstructed patches on the pixel-level measures (L1, PSNR, SSIM). However, what is particularly striking is the consistent boost of $+10\%$ for each method in FID score (with a similar pattern observed for IS). These methods compute a higher-level notion of perceptual similarity by comparing intermediary feature maps from a network pre-trained using a supervised objective for real and generated images, and suggests that through the introduction of a perceptual loss term the decoder learns a more generalisable notion of perceptual similarity. Examples of reconstructed patches are shown in Figure~\ref{fig:recon_focus}. %and Figure~\ref{fig:recon_results}.

\begin{figure*}[t]
    \centering
    \includegraphics[width=1.0\textwidth]{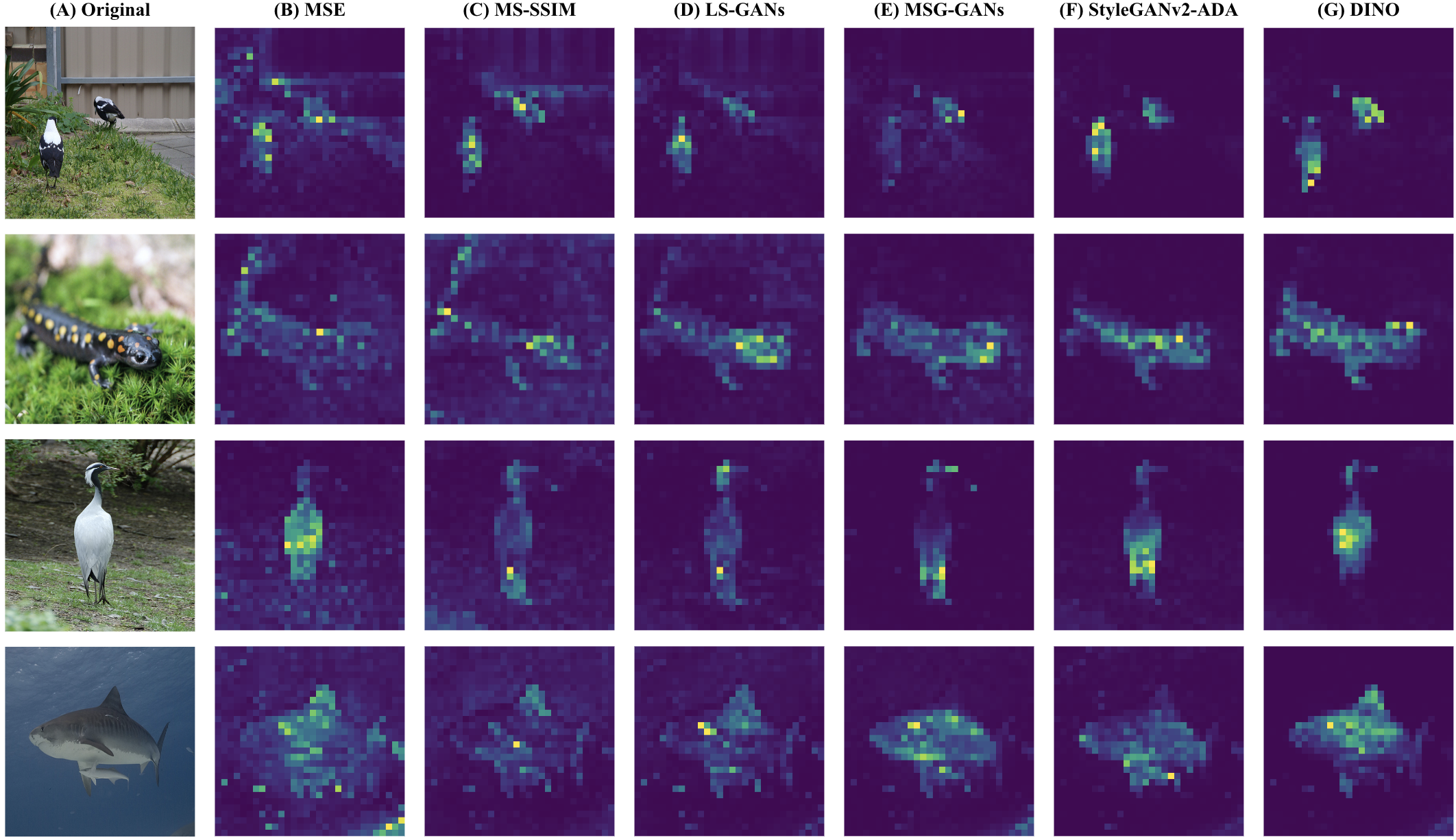}
    \caption{Attention maps of models pre-trained on ImageNet-1K without labels. We visualise the self-attention of the [CLS] token of the last layer. Sample images were selected randomly from the ImageNet-1K validation set. Column (A) is the original input image, (B-F) the outputs from our different losses and (G) is the output from DINO for comparison.} 
    \label{fig:attn_maps}
\end{figure*}

\textbf{Self-attention maps.} To evaluate qualitatively whether the features learnt by our approach properly capture the high-level semantics of the image over low-level details we visualise the attention maps from the final layer of our network, shown in Figure~\ref{fig:attn_maps}. Compared to the original MAE formulation, the combination of perceptual and adversarial loss (LS-GAN) leads to sharper focus on the object in the frame despite no supervision being used during training. The addition of multi-scale gradients (MSG-GAN) and adaptive discriminator augmentation and perceptual path length regularisation (StyleGANv2-ADA) brings further improvement.

In particular, with our best method, we achieve similar qualitative results to DINO~\cite{caron2021emerging} a self-supervised method which takes a contrastive approach to learning and requires careful balancing of the loss and sampling of image crops within batches compared to the simpler reconstruction-based approach used by our method.

\textbf{Linear probing.} To evaluate quantitatively the extent to which the features learnt by our approach capture useful semantic information, a common approach is to freeze the backbone of the pre-trained encoder model and train a simple linear classifier on top. We report the results for this over the IN1K validation set in Table~\ref{tab:imagenet1k}, comparing to the MAE, BEiT and contrastive learning approach MoCo v3.

Our baseline model variant trained via MS-SSIM achieves $71.2\%$ accuracy, $3\%$ higher than the original MAE trained via MSE\cite{MaskedAutoencoders2021}. StyleGANv2-ADA-P attains $\mathbf{78.1\%}$, a boost of $10\%$ compared to the original MAE. This significant increase suggests that our perceptual loss term leads to much more informative features being learnt without fine-tuning with labels being necessary. For comparison, we also include results when using perceptual loss computed instead against a pre-trained network. When adding this stronger supervision, the accuracy further improves to $79.8\%$, although this introduces a dependency on an external network and training data magnitudes larger than IN1K.

\begin{table}
\caption{\textbf{Computational cost.} The relative time to train per epoch. All figures computed with the ViT-B architecture using 8xV100s.}
\vspace{-0.2cm}
  \centering
  \small
  \begin{tabular}{>{\kern-\tabcolsep}cccc<{\kern-\tabcolsep}}
    \toprule
     & MAE & LS-GAN-P & StyleGANv2-ADA-P\\
    \midrule
    % Training time & 15 days & 35 days ($2.3\times$) & 65 days ($4.3\times$)\\
    Rel.\ time / epoch & 0.23 & 0.54 & 1\\
    \# Parameters & 113M & 113M & 119M (+5\%)\\
    \bottomrule
  \end{tabular}
  \label{tab:table_traintime}
  \vspace{-0.5cm}
\end{table}

% \begin{table}
% \caption{\textbf{Computational cost.} The relative time to train per epoch. All figures computed with the ViT-B architecture using 8xV100s.}
% \vspace{-0.2cm}
%   \centering
%   \small
%   \begin{tabular}{>{\kern-\tabcolsep}ccc<{\kern-\tabcolsep}}
%     \toprule
%      & hours & \# parameters\\
%     \midrule
%     MAE & 11.4 (0.23) & 113M\\
%     LS-GAN-P & 24.1 (0.53) & 113M\\
%     StyleGANv2-ADA-P & 48.1 & 119M (+5\%)\\
%     \bottomrule
%   \end{tabular}
%   \label{tab:table_traintime}
%   \vspace{-0.5cm}
% \end{table}

\textbf{Computational properties.} All experiments were run on 8xV100s, and took between 1-3 weeks to train to convergence for ViT-B. The relative training times and parameter counts are shown in Table~\ref{tab:table_traintime}. We also tried training MAE longer (\emph{e.g.}\ 900 epochs instead of 300, to match the total time for StyleGANv2-ADA-P) and did not observe signficiant performance improvement.

\subsection{Downstream Learning Results}
\label{transfer_results}

% We also assess the performance of our pre-trained model when fine-tuned on downstream tasks.

\textbf{Image classification.} We fine-tune our models on IN1K, with the results shown in Table~\ref{tab:imagenet1k}. Using ViT-B, we see a consistent boost in performance across the board by adding a perceptual loss term, obtaining an accuracy of $86.2\%$ with our best method (StyleGANv2-ADA-P) and outperforming MAE and BEiT by $2.6\%$ and $3\%$ respectively.

Moving to ViT-L architecture, we obtain $\mathbf{88.1\%}$ accuracy, outperforming all previous methods training only on IN1K data and in particular resulting in comparable accuracy to that reported by MAE of $87.8\%$ using the much larger ViT-H\textsubscript{448} architecture (632M parameters vs 86M parameters, with input image of size 448 rather than 224).

If we use a pre-trained network with dVAE-P, we obtain a further boost of our best accuracy to $88.6\%$

\textbf{Object detection and semantic segmentation.} For object detection, we fine-tune a Mask R-CNN head on MS-COCO with the results shown in Table~\ref{tab:coco_ade20k}. Our best method achieves $53.5$ AP\textsuperscript{box} using a ViT-B architecture, which outperforms the previous best reported result of MAE trained with ViT-B by $3.2$. Similarly, for semantic segmentation, we fine-tune an UperNet head on ADE20K with the results shown also in Table~\ref{tab:coco_ade20k}. Again using a ViT-B architecture, here we obtain up to $50.4$ mIOU, $1.2$ higher than for MAE.

\textbf{Impact of perceptual loss term.} When training a baseline %adversarial loss
LS-GAN without a perceptual loss term, we are unable to train a model that performs better than the baseline MSE loss~\cite{MaskedAutoencoders2021} and observe clear stability issues during training. However, with an MSG-GAN loss training is much more stable. Referring to Table~\ref{tab:imagenet1k}, for ViT-B with adversarial and reconstruction loss only we obtain a $1.1\%$ boost (a \emph{vs.}\ c) over the baseline MSE loss for image classification. However, this remains less than the $1.7\%$ boost (a \emph{vs.}\ b) with perceptual component added. An even larger gap is observed for ViT-L ($0.6\%$ \emph{vs.}\ $1.3\%$ boost). This suggests the perceptual loss term plays an important role not only for training stability but also is a large driver of performance.

\textbf{Impact of multi-scale MAE. } Training with a multi-scale loss without updating the MAE architecture as described in Section~\ref{msg_mae_model} results in a drop of performance of over $2\%$ for image classification as seen in Table~\ref{tab:imagenet1k} (b \emph{vs.}\ d), with a similar drop also observed for object detection and semantic segmentation in Table~\ref{tab:coco_ade20k} (b \emph{vs.}\ d).

\section{Conclusion}

We explored a method for incorporating the learning of higher-level features from images explicitly into the learning objective of a masked autoencoder. By introducing a perceptual loss term and adversarial training, we showed how the representations learnt by MAE~\cite{MaskedAutoencoders2021} could be improved, boosting transfer performance for downstream tasks such as image classification, object detection and semantic segmentation. In particular, this performance boost is observed not only when fine-tuning, but also in the linear probe setting where contrastive methods have historically done better. This suggests that by combining the rich supervision of the pixel reconstruction task with a more focused higher-level learning signal we can greatly improve the data efficiency of the masked autoencoder approach.

This work also helps to start to address one of the key differences between the use of masked modelling for images and text: that %images and
images and image patches do not have inherent semantic meaning. Many questions remain %though
about how to learn cues of the right level of abstraction directly from image data.
% Future work could further explore how to draw further inspiration from work in the generative modelling literature to this end
% Future work could explore this further,
% combining perceptual understanding from adversarial training with discretised learning of semantic features~\cite{NIPS2017_7a98af17}.

% (i) adversarial learning -> perceptual understanding in our models
% (ii) BEiTv2 -> discrete codebook and combinations, have direct mapping of semantic features (relationships within those images)
% by combining the two -> not only learn semantic feature set, but also learn entirety of distribution embedded in feature maps
% greatly improves visual representation learning
% (a) not only intra-class features, but also inter-class features
% --> intra-class -> codebook determines how eyes relate to nose -> discretising scene
% "clustering of semantic concepts"

%\begin{figure*}[p]
 %   \centering
 %   \includegraphics[width=1.0\textwidth]{image_recon.png}
 %   \caption{Exhibiting random samples from the ImageNet-1K validation set. Where Columns are; (A) the original ground truths, (B) the masked input (MIR 75\%), (C-G) are the reconstructed outputs generated MAE model trained with, MSE\cite{MaskedAutoencoders2021}, SSIM+L1, LS-GANs, MSG-GANs, StyleGANv2-ADA losses respectively.}
 %   \label{fig:recon_results}
%\end{figure*}

%%%%%%%%% REFERENCES
{\small
\bibliographystyle{ieee_fullname}
\bibliography{egbib}
}

\end{document}